\documentclass[10pt,twocolumn,letterpaper]{article}
\usepackage{verbatim}
\usepackage{cvpr}
\usepackage{times}
\usepackage{epsfig}
\usepackage{graphicx}
\usepackage{amsmath}
\usepackage{amssymb}
\usepackage{color}
\usepackage[labelsep=period]{caption}
\usepackage{subfigure}
\usepackage{caption}
\usepackage{cite}
\usepackage[ruled,vlined]{algorithm2e}
\usepackage{multirow}

\usepackage{url}
\usepackage[toc,page]{appendix}
\usepackage{pifont}
\usepackage[pagebackref=true,breaklinks=true,letterpaper=true,colorlinks,bookmarks=false]{hyperref}
\usepackage{array}
\usepackage{enumitem}

\newcommand{\tb}[1]{\textcolor{blue}{#1}}
\newcommand{\tr}[1]{\textcolor{red}{#1}}

\newcommand{\Cgray}[1]{\cellcolor[HTML]{D8D6D6}} %

\def\cvprPaperID{2743} 

\cvprfinalcopy

\begin{document}
	
	\title{Product Inspection Methodology via Deep Learning: An Overview}
    

	\author{Tae-Hyun Kim$^{1}$, Hye-Rin Kim$^{1}$ and Yeong-Jun Cho$^{2}$
		\\\\$^{1}$ Data Science Team, Hyundai Mobis, South Korea
		\\{\tt\small \{th,hyerin.kim\}@mobis.co.kr}
		\\$^{2}$  Department of Artificial Intelligence Convergence, Chonnam National University
		\\{\tt\small yj.cho@jnu.co.kr}
	}
	
	
	\maketitle
	
	\begin{abstract}

		In this work, we present a framework for product quality inspection based on deep learning techniques.
		First, we categorize several deep learning models that can be applied to product inspection systems. Also we explain entire steps for building a deep learning-based inspection system in great detail.
		Second, we address connection schemes that efficiently link the deep learning models to the product inspection systems.
		Finally, we propose an effective method that can maintain and enhance the deep learning models of the product inspection system. It has good system maintenance and stability due to the proposed methods.
		All the proposed methods are integrated in a unified framework and we provide detailed explanations of each proposed method.
		In order to verify the effectiveness of the proposed system, we compared and analyzed the performance of methods in various test scenarios.

	\end{abstract}
	
	
	\section{Introduction}
	
	Many manufacturing companies have applied product inspection systems to detect the product defects and evaluate the quality of the products.
	The inspection system examines the possibility of functional problems of the product or where the defects are existed on the surface of the product.
	To this end, the inspection system generally utilizes several camera sensors to examine all or key parts of the products.
	Some systems that automatically finds the defects of products based on image processing technologies can save human effort and labor.
	
	Unfortunately, many conventional methods for defect detection following a rule-based algorithms have shown the low performances for finding the defects of the products~\cite{putera2010printed,dave2016pcb}.
	For example, the conventional methods are difficult to deal with subtle changes in the environment (\textit{e.g.}, small changes in product location or illumination). 
	In addition, they often fail to detect new types of defects due to their simple criteria --
	if some defective parts are not detected in the current manufacturing step, they will proceed to the consecutive assembly step and lead to significant financial loss.
	Furthermore, field workers should always manage and adjust parameters of the system.
	In that case, First Time Yield~(FTY)\footnote{The number of good units (\textit{i.e.}, products) produced divided by the number of total units going into the process.} of products also decreases.
	
	To overcome the limitations of the rule-based methods, several methods applying deep learning~\cite{wei2018cnn,jing2017yarn, li2018real, li2020rail} have been studied in recent years.
	Deep learning has main advantages and strengths : (1) No need for feature engineering; (2) Large network capacity to learn from low-level feature to high-dimensional representations; (3) Showing superior performance under various conditions (\textit{e.g.},illumination changes, noisy image and etc.).	
	\begin{figure}[t]
		\centering
		\includegraphics[width=1\columnwidth]{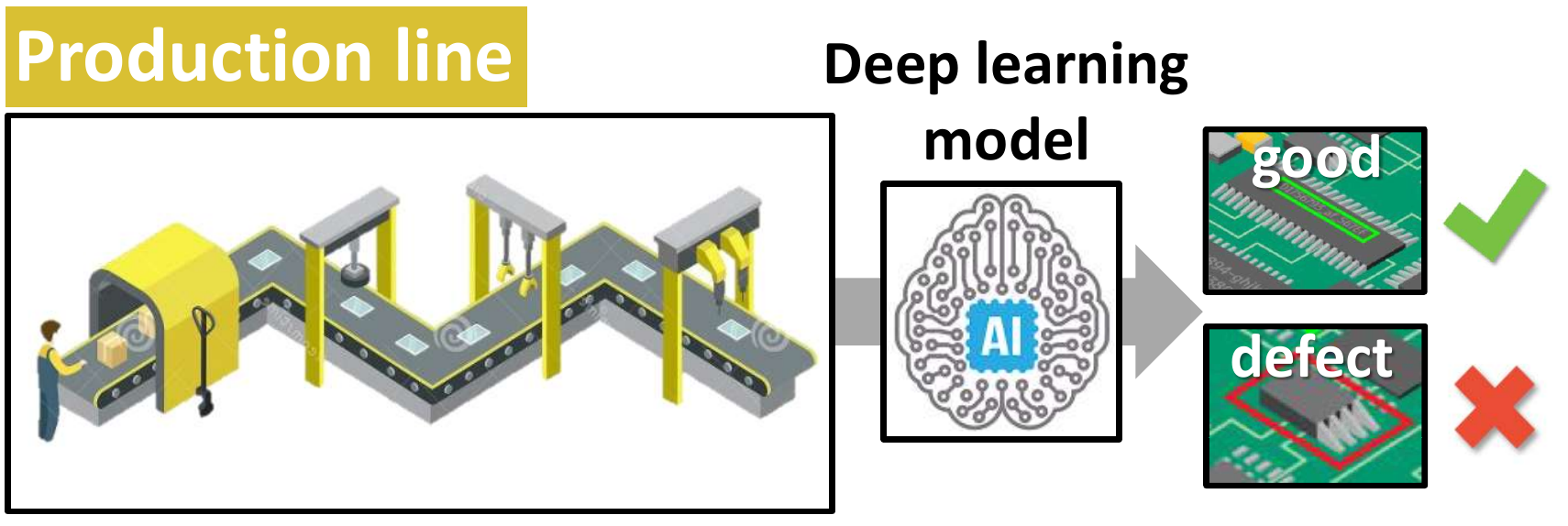}
		\caption{A production line applied deep learning model}
		\label{FIG:intro}
	\end{figure}
	However, in order to adopt deep learning algorithm to a defect inspection system, various issues should be considered as follows:
	\begin{itemize}
		{
			\item A series of steps to train and utilize deep learning models for the product inspection system in details
			\item Choosing proper deep learning models for the system
			\item Connecting deep learning models to existing systems
			\item User interface and maintenance schemes}
	\end{itemize}
	Although there are a lot of issues, there are few works have been studied on how to address the issues and apply deep learning models to the existing product inspection system.	
	
\begin{figure}[t]
		\centering
		\includegraphics[width=1\columnwidth]{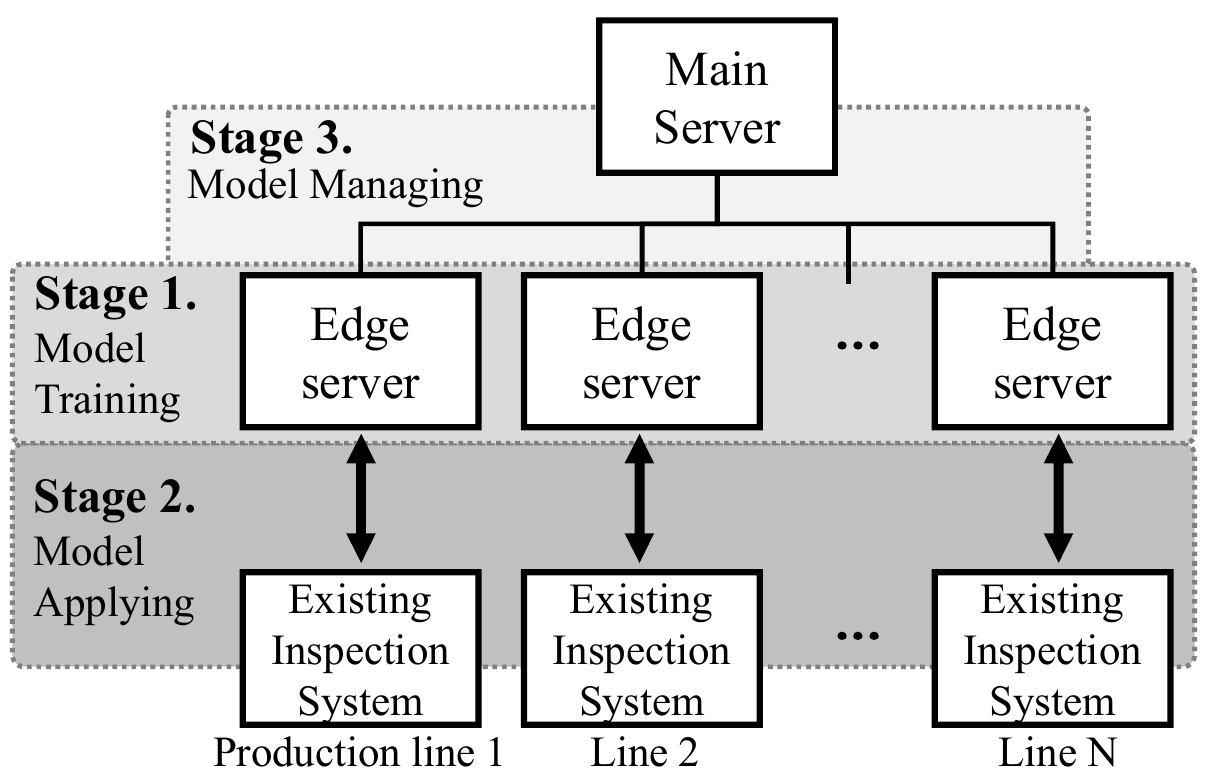}
		\caption{A system hierarchy for automatic product inspection via deep learning and proposing three stages for the systems}
		\label{FIG:whole_frame}
	\end{figure}

	In this work, we efficiently handle all issues based on our knowledge and experience in the real manufacturing field. We aim to build a framework for automatic product inspection based on deep learning techniques as shown in Fig.~\ref{FIG:intro} and the overall proposed system hierarchy is shown in Fig.~\ref{FIG:whole_frame}. Edge servers are assigned for each production line and a main server manages each edge server. There are three main stages for applying deep learning into the product inspection system: 1) Model training stage, 2) Model applying stage, and 3) Model managing stage.
	In model training stage, we explain the entire steps to train deep learning models for the product inspection system in Sec~\ref{sec:train}. We also provides detailed explanations and guidelines for each step including data collection, data pre-processing, choosing proper deep learning models for the system purposes.
	
	In model applying stage, we propose connecting schemes that link the trained deep learning models to the existing inspection system in Sec~\ref{sec:connecting}. In general, many equipment such as product inspection systems in old factories are out of date and their resources are limited. \textit{E.g.}, computing power and storage capacity of them are insufficient. Therefore, it is difficult to operate deep learning models requiring many resources in old systems.
	Instead, we set another workstation as a server for deep learning and connect the server with the existing inspection system.
	As a result, we can operate deep learning models with maximum performance. Moreover, all functions of the existing inspection system such as product managing and control units, inspection equipment~(\textit{e.g.}, jigs, cameras, lights and Etc.) and software can be utilized very stably. Once we successfully train and apply a deep learning model; then we recycle and expand the trained model to another production lines as described in Sec.~\ref{subsubsec:ME}. This model expansion method reduces human effort for training additional deep learning models.
	
	Finally, in model managing stage, we propose an effective model update method that can maintain and enhance the deep learning models of the product inspection system. We first employed Grad-cam~\cite{selvaraju2017grad} providing visual explanations from deep networks helps system managers to understand predictions of deep learning~(Sec.~\ref{subsec:exp_system}). The proposed model update method includes model \textit{fine-tuning} and \textit{re-training} processes as described in Sec.~\ref{subsec:model_update}. The proposed system is fully automated. Moreover, it shows outstanding system maintenance performance and stability. 
	Thanks to the proposed system, even the system managers who do not understand the deep learning techniques can manage the system very easily.
	
	In order to verify the effectiveness of the all proposed system, 
	we compared and analyzed the performance of methods in various test scenarios.
	As a result, we can build an automatic product inspection system based on deep learning in a unified framework.
	To sum up, the main contributions of this paper can be summarized as follows: (1) The first attempt to provide whole steps for building a production quality inspection system based on deep learning including many helpful guidelines and technical know-how, (2) Proposing very practical and effective model update methods for the system maintenance, (3) Intensive verification of proposed methods and comparison with other conventional methods. 
	
	We hope that our work is helpful and would provide useful guidelines to readers who desire to implement deep learning-based system for the product inspection.

	\section{Related works}
	\label{sec_rw}
	
	\subsection{Conventional Defect Inspection Systems}
	
	When visual inspections are conducted by humans, inspection error may occurred due to human factors such as fatigue, inconsistency, inability to unify the test criteria, Etc.
	In order to reduce human error, many simple methods~\cite{putera2010printed,dave2016pcb,soini2001machine} have been studied to perform product inspection that automatically finds the location of the defects or classifies the types of defects of the products. They employ simple image processing technologies such as image thresholding and binarization~\cite{otsu1979threshold}.
	Unfortunately, the methods are too simple to deal with subtle changes in the environment (\textit{e.g.}, small changes in product location or illumination) and show low inspection performances.

	\begin{figure*}[t]
		\centering
		\includegraphics[width=1.9\columnwidth]{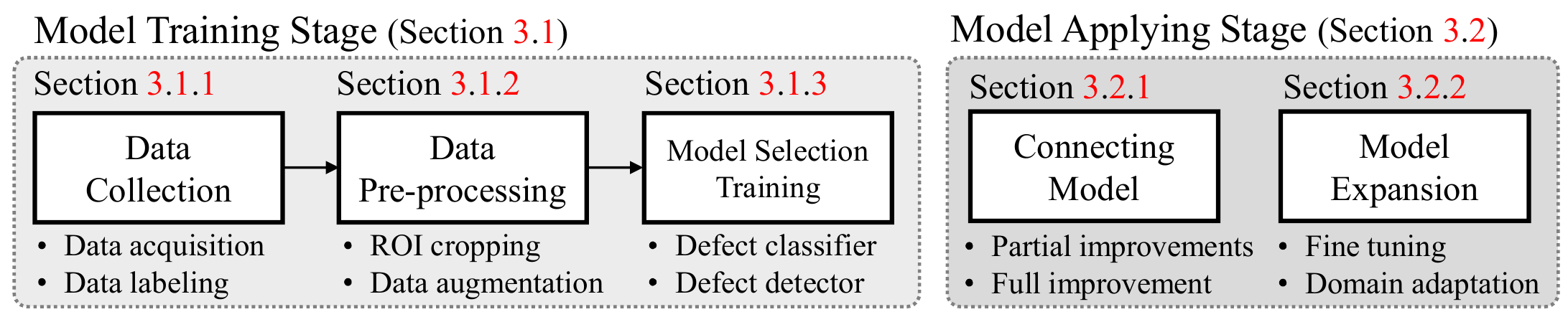}
		\caption{Deep learning model training and applying stages for a defect inspection system}
		\label{FIG:framework}
	\end{figure*}
	
	In order to overcome the limitations of the simple methods, Chang~\etal~\cite{chang2008case} proposed more sophisticated method so-called two-phase methodology.
	In addition, Chaudhary~\etal~\cite{chaudhary2017automatic} employed machine learning techniques to propose the high-performance defect detection system. 
	Although machine learning techniques have shown positive results in improving inspection system performance, it is still not easy to manually grasp features which are suitable for defect detection and generate a new model each time in a situation where new types of defects are constantly occurring.

	\subsection{Defect Inspection with Deep Learning}
	
	With the recent development of the deep learning, many attempts to adopt the deep learning into the product inspection system are highly active.
	In particular, not only object recognition and classification studies, which are conducted to determine the quality of products based on product image data, but also the defect detection technique studies, which find the location of product defect occurrence, are ongoing.
	
	Studies related to object recognition and classification include research using AlexNet~\cite{krizhevsky2012imagenet} to recognize defects in dyed fibers or fabrics~\cite{jing2017yarn}. Additionally, there is a study~\cite{li2020rail} that performs defect detection using sliding window methods to distinguish the poor surface conditions (such as scratch, poor junction, etc.). 
	Furthermore, in case of finding the locations of the defects, there are studies~\cite{li2018real, wang2018surface} on detecting defects on the surface of steel and in the paper paltes based on yolo-series~\cite{redmon2018yolov3,bochkovskiy2020yolov4} and Mask R-CNN~\cite{girshick2014rich} respectively. 
	Several works~\cite{li2018real,wang2018surface} compared performances of state-of-the-art deep learning algorithms for product inspections.
	In addition to building an algorithm or model for product inspection, it is also necessary to consider how to select a deep learning model that fits the characteristics of the product, how to connect the deep learning model to the existing system and how to manage the new system. However, there are few works have been studied on how to address the issues and apply deep learning models to the existing system. In this work, we propose a new framework that considers all of these issues.
	

	\section{Proposed Automatic Product Inspection \\ Framework via Deep Learning}
	\label{sec:system}
	
	To check the qualities of products, the systems such as automatic optical inspection (AOI) system and vision inspection system have been introduced in the field of manufactoring. 
	In general, these systems utilize visual sensors such as RGB cameras or infrared cameras with various illuminating conditions to examine the key parts of the products.
	
	In this section, we explain the overall process for the proposed product inspection system.
	We categorize the proposed process into three main stages: (1) deep learning model training stage in Sec.~\ref{sec:train}; (2) deep learning model applying stage in Sec~\ref{sec:apply_stage}; deep learning model managing stage in Sec.~\ref{sec:managing_stage}. 
	The overall processes of first and second stages are summarized in Fig.~\ref{FIG:framework}.
	
	\subsection{Model Training Stage}
	\label{sec:train}
	
	\subsubsection{Data collection}
	\label{sec:data_collection}

	\noindent $\bullet$ \textbf{Image data acquisition.}
	In general, building a deep learning model requires a lot of image data. In case of an inspection system, however, the quantity of the data which can be collected depends on various manufacturing conditions such as production volume, period and data storage. Because data is generated only when a product is manufactured, it is difficult to create as much data as we want. Thus the quantity of data could be limited. In addition, as images generated by the inspection system generally need a huge storage space, there is also a limit in the number of images that can be stored.
	Note that it should be checked how much data storage space is left and make sure to back up date or add more storage space so that the data will not be deleted.

	\noindent $\bullet$ \textbf{Data labeling.}
	After data acquisition step, the collected data should be labeled as `1'~(OK:~non-defective) or `0'~(NG:~defective) according to its surface condition.
	We denote the collected images as $\mathbf{x}_{i}$ where $i$ is an index of the image and the labels of $i$-th images as $y_i$.			
	As the quality and reliability of training data have a substantial impact on the performance of the deep learning model, labels of data should be decided very carefully. 	
	To this end, many experts such as product quality engineers and deep learning engineers need to collaborate with each other. 
	Potential defects of products are classified in advance according to their types. Also, evaluation methodologies and policies are clearly established.
	Then, the collected images are labeled according to predefined rules for building a training dataset as
	\begin{equation}
		\mathcal{D}=  \big\{ \big(\mathbf{x}_i,y_i\big)  | y_i \in  \big\{0,1\big\} , i=1,...,N\big\},
	\end{equation}
	where $N$ is the total number of collected images.
	
	The steps mentioned above will not only increase the reliability of models, but also give clear guidelines to system engineers.
	E.g., when an unseen type of defect occurs in the future, we can easily categorize the defect based on the policies.

	\begin{figure}[t]
		\centering
		\subfigure[Raw image]{\includegraphics[height=0.3\columnwidth]{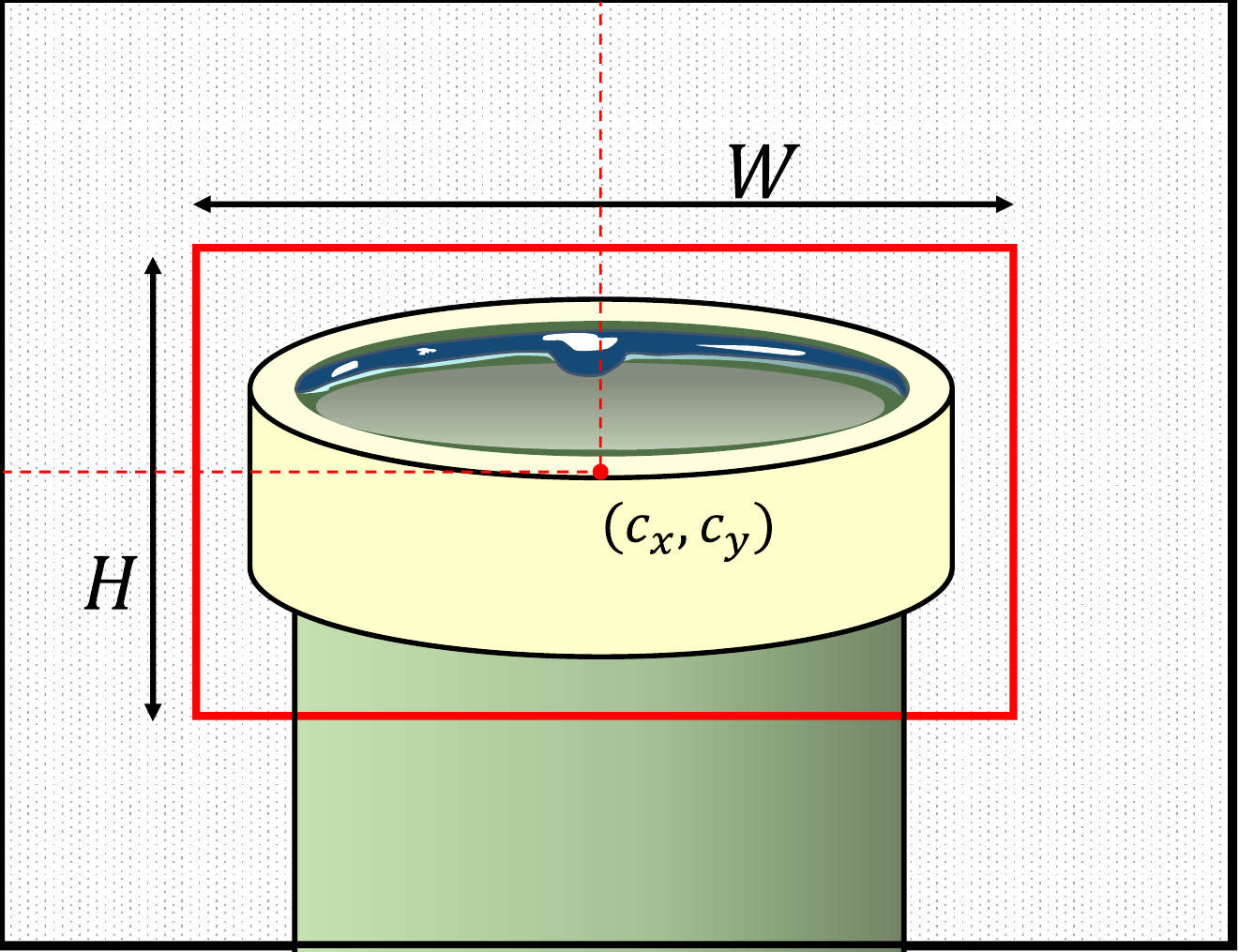}}
		\hspace{10pt}
		\subfigure[Cropped image]{\includegraphics[height=0.3\columnwidth]{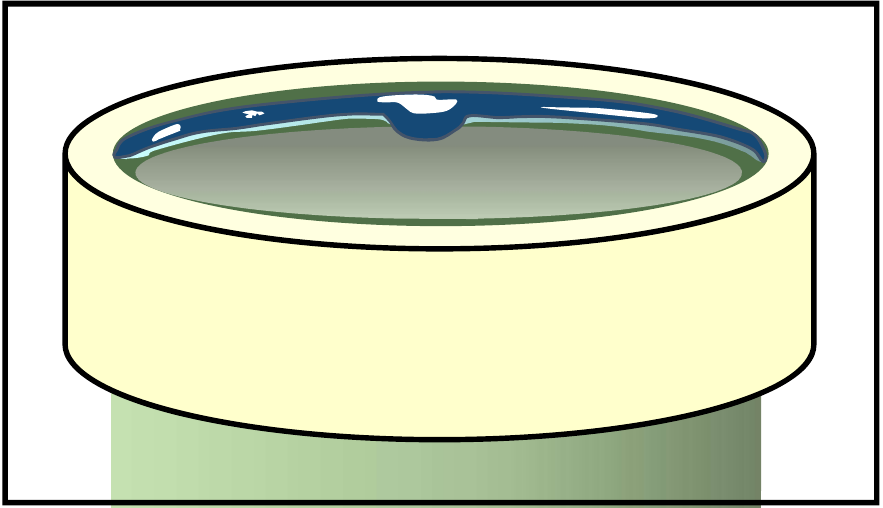}}
		\caption{Example of ROI selection: (a) Raw image: shaded area denotes a background region, solid line denotes a product, red box denotes ROI. (b) Cropped image according to ROI.}
		\label{FIG:crop}
	\end{figure}

	\subsubsection{Data Pre-processing}
	\label{sec:ss2}
	
	Initially, images acquired from the inspection system are usually not optimized and not sufficient to train deep learning models.
	In this section, we discuss data pre-processing that refines and augments image data for training deep learning models.
	
	\noindent $\bullet$ \textbf{ROI cropping}
	Many defect inspection systems contain conveyor belts to transport products and also contain lots of jigs for fixing products to examine their qualities.
	In general, they consistently fix the products and capture the product images. 
	We call the initial product image as a raw image.
	However, the raw image may include an unnecessary background depending on the inspection system as shown in Fig.~\ref{FIG:crop}~(a).
	In this case, we need to set a region of interest (ROI) of the manufactured product. ROI denotes the boundaries of the product on the raw image. We usually set a center position ($c_x$, $c_y$) and the size ($H$, $W$) of the product.
	Then, ROI can be represented as $\mathbf{b}=\big\{c_x,c_y,H,W\big\}$.

	Specifying the locations of products (\textit{i.e.} ROI) is simple but very effective: it rejects lots of unnecessary regions and makes the analysis methods (Sec.~\ref{sec:ss3}) to focus entirely on the products as shown in Fig.~\ref{FIG:crop}~(b).
	In order to set ROIs, it depends on the product fixing and localizing accuracy of the inspection system as follows:
	When products are always located in the same region in the image, we can strictly set ROIs of the products.
	On the other hand, when the product localization is unstable, we have to set ROIs with loose range so that the cropped image based on the ROIs does not miss the product. Otherwise, one possible solution is to apply object detecting methods such as Yolo v3~\cite{redmon2018yolov3} to set the ROIs under the unstable product position in the raw image.
	Please refer to several ROIs of a product in Fig.~\ref{FIG:crop}~(a).

	\noindent $\bullet$ \textbf{Data augmentation}
	As we mentioned in Sec.~\ref{sec:data_collection}, the amount of collected data depends on various manufacturing conditions. Therefore a sufficient amount of data for training deep learning models may not be ensured. Unfortunately, deep neural network commonly requires huge amount of training data for learning lots of network parameters (\textit{i.e.}, weights and bias).
	To deal with the lack of data and to make the training dataset covering various data distributions, we can introduce data augmentation techniques that generates new training data from existing data.
	We summarized several possible data augmentation methods in Table~\ref{Tab:image_aug}.
	There are two categories of data augmentation methods as follows:
	\begin{itemize}
		\item \textit{Image transformation} which is a traditional data augmentation method that changes or transforms the given images to augment new image data by following methods: (1) Perspective transformation changes an image in terms of its size, rotation and perspective. It has 8 degrees of freedom~(DOF) and transforms the image as if the camera observed the images from different viewpoints.
		(2) Color transformation changes color distribution of images and color space~(e.g., RGB, HSV, Etc.) of images.
		(3) Noise addition adds various kinds of noise onto images such as salt-and-pepper, Gaussian and Poisson noise. We can add those noise by using kernel filtering.
		\item \textit{Image generation} creates new images based on the distribution of acquired images. Generative adversarial network~(GAN)~\cite{goodfellow2014generative} can be used.
	\end{itemize}
	Shorten \etal \cite{shorten2019survey} describes image data augmentation for deep learning. 
	In Sec.~\ref{sec:da}, we show some image classification results before and after data augmentation.

	\begin{table}[t]
		\caption{Methods of image data augmentation~\cite{shorten2019survey}}
		\centering
		{\small
			\begin{tabular}{c|l}
				\hline
				\noalign{\hrule height 1pt}
				\textbf{Category}       & \textbf{Method}                \\ \noalign{\hrule height 1pt}
				\multirow{3}{*}{\begin{tabular}[c]{@{}c@{}}Image\\ transformation\end{tabular}} & Projective transformation      \\ \cline{2-2} 
				& Color transformation           \\ \cline{2-2} 
				& Noise addition                 \\ \hline
				\begin{tabular}[c]{@{}c@{}}Image\\ generation\end{tabular}                      & Generative Adversarial Network~\cite{goodfellow2014generative} \\
				\noalign{\hrule height 1pt}
		\end{tabular}}
		\label{Tab:image_aug}
	\end{table}

	\begin{figure*}[t]
		\centering
		\includegraphics[width=1.8\columnwidth]{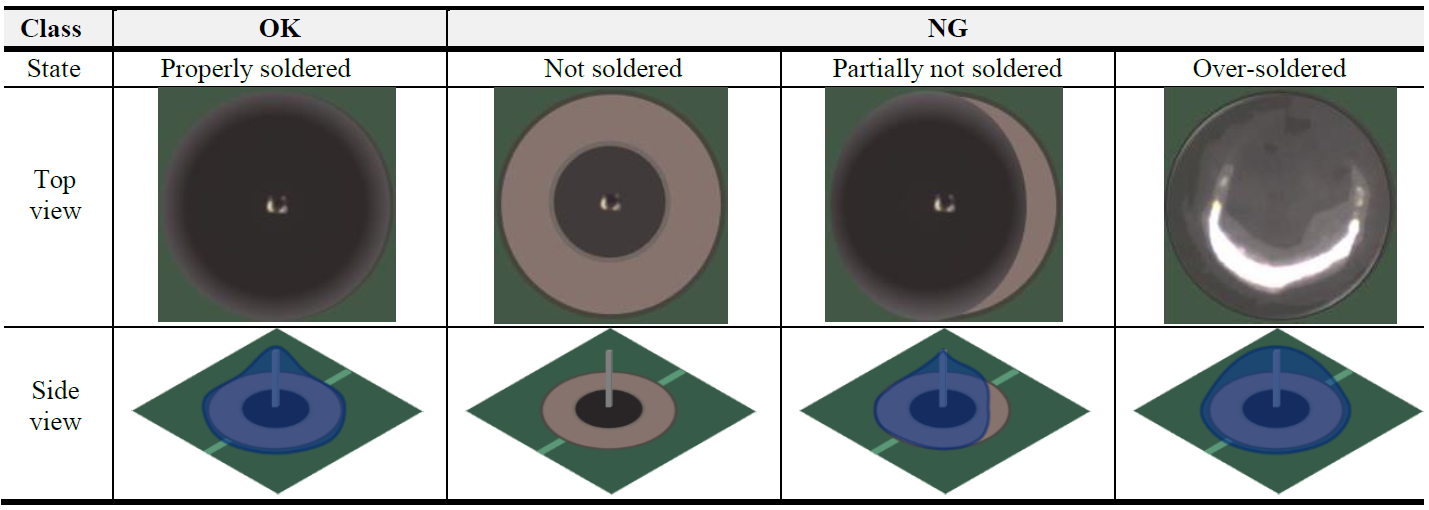}
		\caption{Examples of non-defective~(OK) and defective~(NG) soldered pin. The blue area of side view shows the result of soldering. Defects appear in the overall area of the product part.}
		\label{FIG:ok_ng_images}
	\end{figure*}

	\begin{figure}[t]	
		\centering
		\subfigure[Side view: red shaped area denotes ROI. The system finds defects in the ROI.]{\includegraphics[height=0.43\columnwidth]{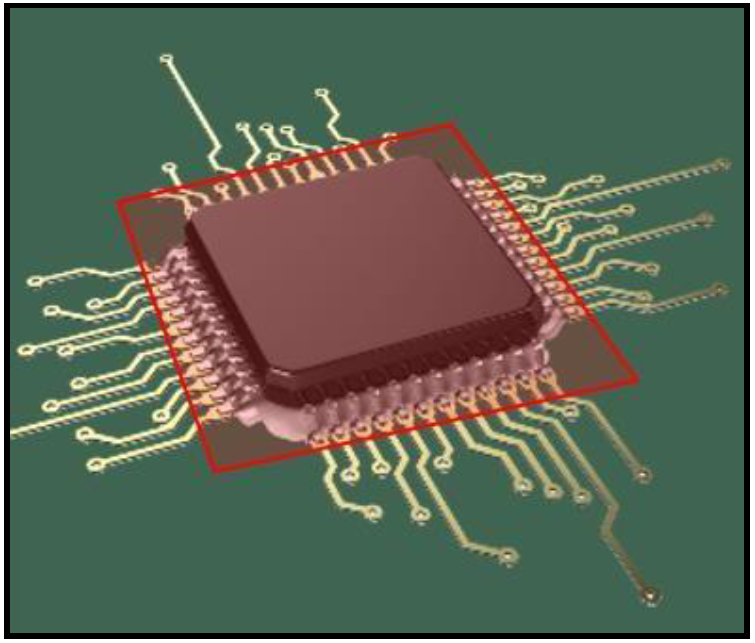}}
		\hspace{0pt}
		\subfigure[Top view: red boxes denote detected defects: abnormal short and hair. ]{\includegraphics[height=0.43\columnwidth]{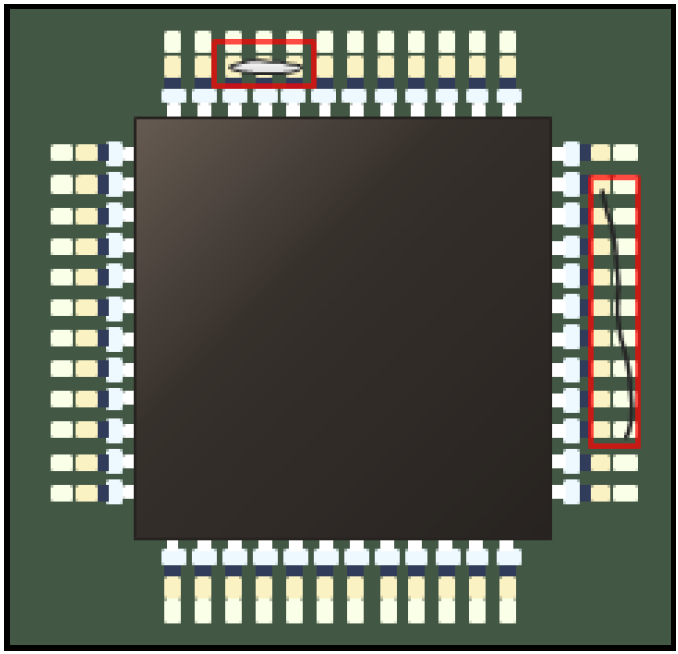}}
		\caption{Examples of partial defects on a product}
		\label{FIG:ok_ng_images_mnt}
	\end{figure}

	\subsubsection{Model Selection \& Training: \\ Classifier vs. Detector}
	\label{sec:ss3}

	We expect that the training dataset $\mathcal{D}$ for learning deep neural networks is prepared through previous Sections (Sec.~\ref{sec:data_collection} and~\ref{sec:ss2}). Now, it is necessary to select a proper deep learning model to build the defect inspection system. 
	In general, the types of products and defects inspected by the defect inspection system are very diverse and different types of inspection methods are required for each of them.
	In this work, we categorize the deep learning models into two types: \textit{defect classifier} and \textit{defect detector}.
	In this section, we discuss how to select proper deep learning models according to characteristics of defects on products.
	
	A common case of product defects is that they appears in the overall area of the product.
	Figure~\ref{FIG:ok_ng_images} shows examples of non-defective~(OK:1) and defective~(NG:0) soldered pins.
	As you can see, the exteriors of defective soldered pins~(NG) have been modified or the lead is completely missing. 
	In such cases, defect classifiers are effective.
	They compare the distribution of two classes~(1,0) and train optimal classifiers that distinguish non-defective parts and defective parts.
	Based on a huge amount of training dataset $\mathcal{D}$, we can train weights~($\mathbf{w}_c$) of a defect classifier. The defect classifier can be represented by a conditional probability distribution as
	\begin{equation}
		p\big(y|\mathbf{x};\mathbf{w}_c\big), 
		\label{eq:class}
	\end{equation}
	where $\mathbf{x}$ is an input image and $y$ is a predicted label of the given image $\mathbf{x}$.
	The classifier $p\big(y|\mathbf{x};\mathbf{w}_c\big)$ predicts the label of given image and it lies on $[0,1]$. We can decide the predicted labels of sample according to probability as follows
	\begin{equation}
		{ y }^{ pred }= \begin{cases} 1 \quad \text{if}~ p\big(y|\mathbf{x};\mathbf{w}_c\big) \geq 0.5 \\ 0\quad \text{otherwise} \end{cases}.
		\label{eq:label}
	\end{equation}	
	We can utilize several deep neural network models such as ResNet~\cite{he2016deep}, googleNet~\cite{szegedy2015going}, and VGGNet~\cite{simonyan2014very} as a backbone network for defect classifiers.
	
	\begin{table*}[t]
		\centering
		\caption{Differences between classification and detection}
		{\small
			\begin{tabular}{r|l|l}
				\noalign{\hrule height 1pt}
				& \multicolumn{1}{l|}{\textbf{Defect classifiers}} & \textbf{Defect detectors} \\ \noalign{\hrule height 1pt}
				Type of defect    & Overall and various kinds of defect      &  Local and similar forms     \\ \hline
				Output    & Predict a class (OK/NG) of the image      & Detect defects in the images     \\ \hline
				References  & ResNet~\cite{he2016deep}, GoogleNet~\cite{szegedy2015going}, VggNet~\cite{simonyan2014very}, alexNet~\cite{krizhevsky2012imagenet}      & Yolo v3~\cite{redmon2018yolov3}, Yolo v4~\cite{bochkovskiy2020yolov4}, EfficientDet~\cite{tan2020efficientdet} \\ \hline
				\noalign{\hrule height 1pt}
		\end{tabular}}
		\label{Tab:class_det}
	\end{table*}
	
	On the contrary, many defects would occur partially or locally in the product's ROI (Fig.~\ref{FIG:ok_ng_images_mnt}~(a)).
	For example, foreign matters (e.g., dust, hair and pollutants) on the products and abnormal short circuits are considered as defects. Unfortunately, the locations of them are not specified but they occur partially or locally on the product as shown in Fig~\ref{FIG:ok_ng_images_mnt}~(b).
	The actual defect only accounts for a very small portion of the product's ROI; therefore the defect classifier that compares overall area of products to classify non-defective or defective are not effective.
	
	In this case, we can employ defect detectors that directly detect the location of defects on products. In general, the types of foreign matters are not diverse; therefore we can easily specify the defects~(e.g., dust, hair, pollutants and abnormal short circuits) to train defect detectors.
	By using the training dataset $\mathcal{D}$, we train weights ($\mathbf{w}_d$) of a defect detector. Note that for training the defect detector, defect positions $(\mathbf{b}=\big\{c_x,c_y,H,W\big\})$ of training samples in $\mathcal{D}$ should be prepared in advance. Refer to Sec.~\ref{sec:ss2} and check ROI cropping processes which are exactly the same with setting defect positions of each sample.
	The defect detector can be represented by a conditional probability distribution as
	\begin{equation}
		p\big(\mathbf{b}|\mathbf{x};\mathbf{w}_d\big),
	\end{equation}	
	where $\mathbf{x}$ is an input image and $\mathbf{b}$ is a predicted position of a defect.
	The detector $p\big(\mathbf{b}|\mathbf{x};\mathbf{w}_d\big)$ predicts defect positions of given image and it lies on $[0,1]$. In general, if the probability of the detector is larger than `$0.8$', we can decide a defect occur at the position $\mathbf{b}$.
	When any defect is found, the product (\textit{i.e.,} sample image $\mathbf{x}$) is considered defective.
	For the defect detectors, Yolo v3, v4~\cite{redmon2018yolov3,bochkovskiy2020yolov4} and efficient-Det~\cite{tan2020efficientdet} show superior performance in terms of both detection accuracy and speed.	
	As we mentioned above, defect detectors that directly find the location of defects are very effective where their are partial defects on products.
	In Sec.~\ref{subsecar 4.1}, we show the effectiveness of defect detectors.

	\begin{figure*}[t]
		\centering
		\includegraphics[width=1.9\columnwidth]{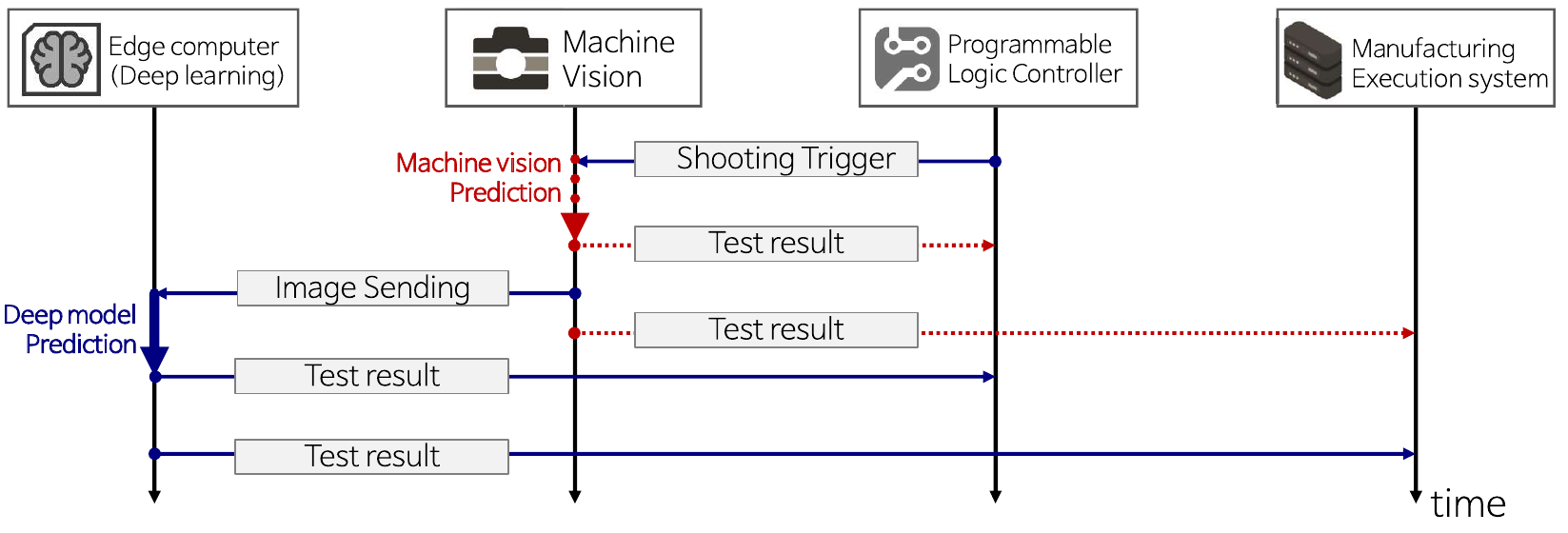}
		\caption{System flowchart to improvements of the existing system based on a deep learning. In order to achieve partial improvements~(goal 1, 2) whole processes~(\tb{blue arrows} and \tr{red-dotted arrows}) in the flowchart are required. For achieving complete replacement of the old system by deep learning~(goal 3), only several processes~(\tb{blue arrows}) are required.}
		\label{FIG:system1}
	\end{figure*}
	
	Note that we have to consider not only accuracy, but also operating speeds of models.
	Not always, but in general, accuracy and operating speed of the deep learning model follow a trade-off relationship. For example, a model performs a good inspection accuracy but it requires too much operating time. Then the model cannot be applied to the inspection system, because products should be manufactured within cycle time.
	The trade-off must be considered before the application of the deep learning models.
	We summarized some practical models for defect classification and detection in Table~\ref{Tab:class_det}.

	\subsection{Model Applying Stage}
	\label{sec:apply_stage}

	\subsubsection{Connecting Deep Learning Models \\ to Existing Systems}
	\label{sec:connecting}
	
	Many product inspection systems in old factories are out of date and their resources such as computing power and storage capacity are insufficient. Therefore, it is difficult to operate deep learning models requiring lots of resources in old systems.
	Instead, a possible simple solution is to set another workstation as a server for operating deep learning models and connect the server with the existing inspection system. 
	Since we have separated each system, deep learning model can operate with maximum performance.
	Moreover, all functions of the existing system such as product managing software, control units and inspection system including equipment~(\textit{e.g.}, jigs, cameras, lights and Etc.) can be utilized very stably.
	In this work, we call the workstation running deep learning models the \textit{edge computer}.
	
	According to the current drawback of the existing inspection system, we have to clearly define the application purpose of deep learning and the improvement goal of the existing system.
	The application steps are categorized in three different goals as follows:

	\begin{table*}[]
		\centering
		\caption{Explanation on terminologies}
		\label{tab:terminology}
		{\small
			\begin{tabular}{r|l}
				\noalign{\hrule height 1pt}
				\textbf{Terminology}   & \textbf{Description}                                                                                              \\  
				\noalign{\hrule height 1pt}
				Shooting trigger & Requesting the machine vision to start the inspection  \\ \hline
				Image sending & Sending images taken by machine vision to edge computer \\ \hline
				Machine vision prediction & Product inspection by machine vision \\ \hline
				Deep model prediction &  Product inspection by deep learning model \\ \hline
				Test result &  Inspection result whether the product non-defective~(OK) or defective~(NG)  \\ \hline                                                                                                          \noalign{\hrule height 1pt}
		\end{tabular}}
	\end{table*}

	\begin{enumerate}
		
		\item Reducing false positive rates of the existing system
		\item Improving true positive rates of the existing system
		\item {\small Replacing the existing system with deep learning models}
	\end{enumerate}
	
	When false defects~(\textit{i.e.,} false positive rates) occur frequently in the existing system, a deep learning model that prioritizes reducing false defects can be applied. 
	It improves the worker productivity and product yield: the workers do not need to perform unnecessary product re-inspection.
	On the other hand, when the defect detection rate~(\textit{i.e.,} true positive rates) of the existing system is poor, this also can be improved with deep learning models for enhancing the detection rate. 
	Therefore the product stability and reliability are improved.
	To sum up, goal 1 and goal 2 are partial improvements of the system.
	When both goals are achieved by the trained deep learning model, then a complete replacement of the existing inspection system~(goal 3) by deep learning will naturally follow.
	After setting the improvement goals, then the connection between the edge computer and an existing system is needed.	
	In general, the product inspection system mainly consists of three manufacturing parts: 1) Machine Vision Inspection (an existing inspection system), 2) Manufacturing Execution System~(MES), 3) Programmable Logic Controller~(PLC)\footnote{Please refer to Appendix.~\ref{sec:MES_PLC} for the details of MES and PLC.}.
	According to your improvements goals, connection schemes are divided into two ways.
	
	First, when you aim to partially improve the old system (goal 1 and 2), then a connection scheme follows the flowchart in Fig.~\ref{FIG:system1}.
	To reduce false positive rates~(goal 1), edge computer with deep learning model only predicts the samples confirmed \textit{defective} by the machine vision inspection.
	\textit{Non-defective} samples confirmed by the machine vision are sent to PLC and MES.
	Similarly, to improve true positive rates~(goal 2), the edge computer only predicts the samples confirmed \textit{non-defective} by the vision inspection. 
	\textit{Defective} samples confirmed by the machine vision are sent to PLC and MES.
	Note that each goal aims to enhance the drawbacks of existing machine vision systems.
	When label predictions are done by edge computer, the edge computer directly sends the inspection results to PLC and MES.
	
	Second, for the complete replacement of the old machine vision system~(goal 3), the connection scheme only requires blue arrows of the flowchart in Fig.~\ref{FIG:system1}.
	Compared to goal 1 and 2, it is much simpler because it does not consider the prediction results by old machine vision system --	The machine vision does not carry out inspection but only records product images to send them to edge computer. Then, the deep learning model inspects the products and send the results to PLC and MES.
	To communicate between different systems, we designed a simple sockets program in C\# language based on TCP/IP protocols.
	All terminologies on the flowcharts are summarized in Table~\ref{tab:terminology}.
	
	Even when a deep learning model is successfully connected to the old system, it cannot be applied to production immediately.
	At least a month of validation period is required to confirm the stability and accuracy of the system and the model. During the period, it should be monitored consistently whether there has been any data bottleneck or abnormal shutdown in the system.

	\subsubsection{Model Expansion}
	\label{subsubsec:ME}
	
	If we successfully trained a deep learning model for a specific product inspection system, we can consider 'model expansion' to improve deep learning models at other inspection systems. Generally, mass production plants produce similar types of products in parallel at different lines. However, if we want to train the deep learning models of each production line separately, we need to collect a large amount of data for each inspection system, which requires a great deal of time and manpower.  
	
	To address these challenges, we can first leverage simple transfer learning technique, fine-tuning~\cite{pan2009survey}. Fine-tuning is one of the methods to reduce the number of required data for learning target inspection models while expanding deep learning inspection systems to other lines. We first set initial values of the deep-learning network with parameters of a trained model, which is already used in another line, and then perform additional training and parameter modification with the training data obtained from a target inspection system. In this case, it has the advantage of ensuring a certain level of performance, even if there is only a small amount of training data because training is not started from random parameters. However, if the difference between defect types of each line is considerable or there are many environmental gaps, such as illumination or location of parts, it can be difficult to expect great effects with simple fine-tuning.
	
	Domain adaptation can be used to overcome such challenges~\cite{ben2007analysis}. This technique adapts two different domain distributions to reduce the discrepancy. This allows the domain distribution of target lines to be adapted to the distribution of a pretrained deep learning model which is already used in different lines. Specifically, in the research introduced in~\cite{kim2020a}, domain adaptation technique is effectively used for model expansion at real-world manufacturing lines.

	\subsection{Model Managing Stage}
	\label{sec:managing_stage}
	
	It is highly likely that system managers lack understanding of artificial intelligence and deep learning technologies.
	Since deep learning models are generally not easy to modify intuitively, it is difficult for the system managers to maintain and supplement deep learning models.
	While the conventional product inspection systems can be easily maintained by adjusting few system parameters, a deep learning model requires much more complex tasks. The system managers need to retrain the deep learning model regularly to make the model robust to unseen data. In addition, they should check whether the proper re-training process has been carried out.
	In this work, we propose deep learning model management system for more easier and flexible system maintenance.

	\begin{figure}[t]	
		\centering
		\includegraphics[width=1\columnwidth]{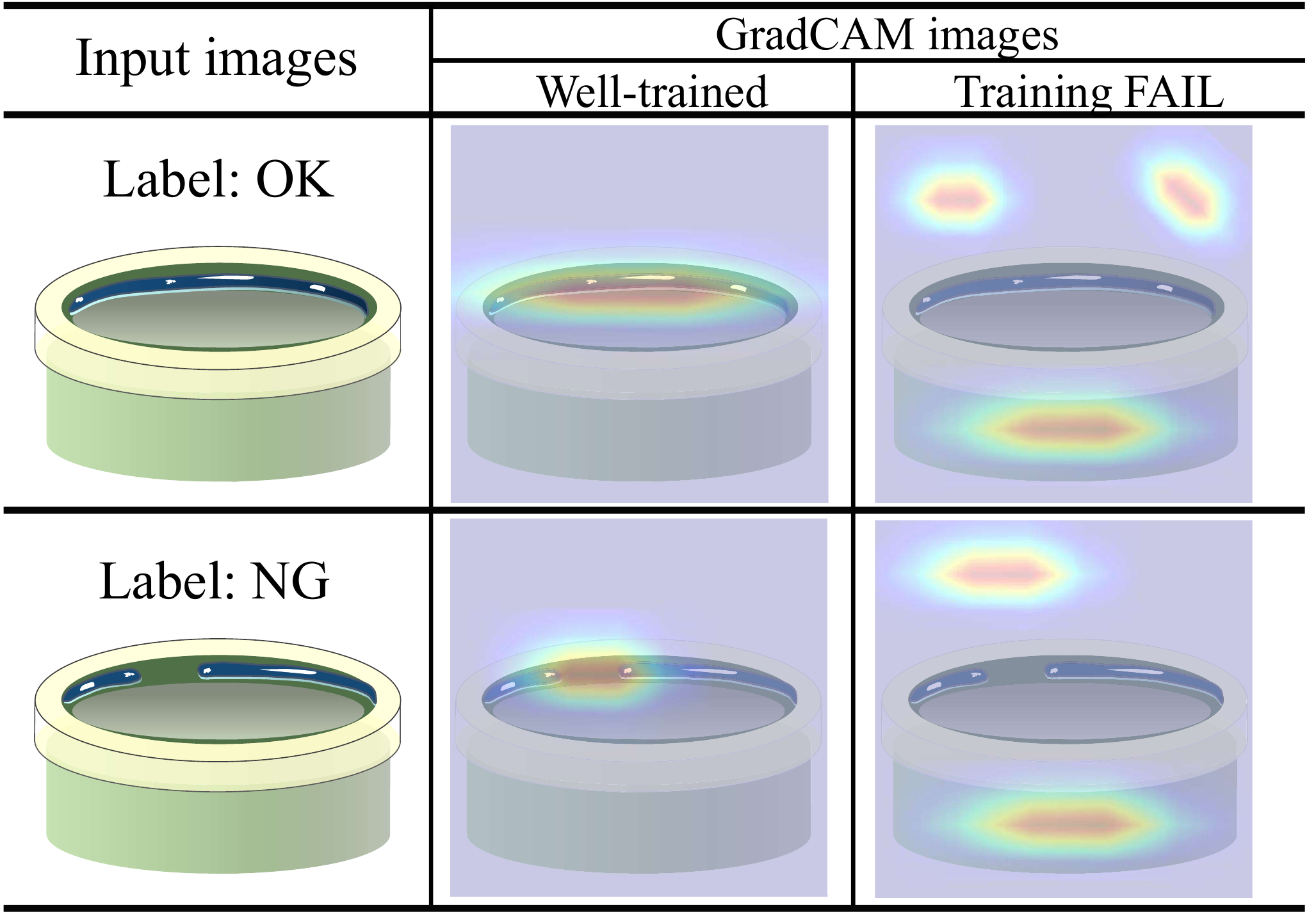}
		\caption{Grad-cam images of cylinder bonding images. Warm color indicates high importance.}
		\label{FIG:gradCAM}
	\end{figure}

	\subsubsection{Explainable System: Grad-cam}
	\label{subsec:exp_system}
	
	System managers who do not have any background knowledge of deep learning are hard to manage the product inspection system based on deep learning.
	The biggest challenge for them is that they cannot judge whether the deep model is properly trained or not.
	In this work, we exploit Grad-cam~\cite{selvaraju2017grad} to alleviate the challenge and utilize the results of Grad-cam to build efficient model update system.
	
	Grad-cam~\cite{selvaraju2017grad} provides visual explanations from deep networks and highlights the important regions in the image for predicting the labels.
	Therefore, managers with no deep learning knowledge can easily analyze the results of the Grad-cam.
	Figure~\ref{FIG:gradCAM} shows examples of non-defective~(OK) and defective~(NG) cylinder bonding images with Grad-cam results. Bond should be evenly covered inside the cylinder -- If the bond is broken or lumped together, it is defective.
	When the deep networks trained well, the Grad-cam focuses on important regions~(i.e., bonding regions) for classifying whether the product defective or not.
	Otherwise, Grad-cam focuses on unimportant regions such as background~(see Training FAIL cases in Fig.~\ref{FIG:gradCAM}).
	
	Although the deep model succeed to classify the labels, if the Grad-Cam result of the sample is not reliable, the model cannot cover the sample yet.
	To make the deep model more robust, system managers collect the image samples with unreliable Grad-Cam and re-train the deep network model with the samples.
	We explain the proposed model update system in the following Sec.~\ref{subsec:model_update}.
	
	\begin{figure}[t]	
		\centering
		\subfigure[label verification]{\includegraphics[height=0.32\columnwidth]{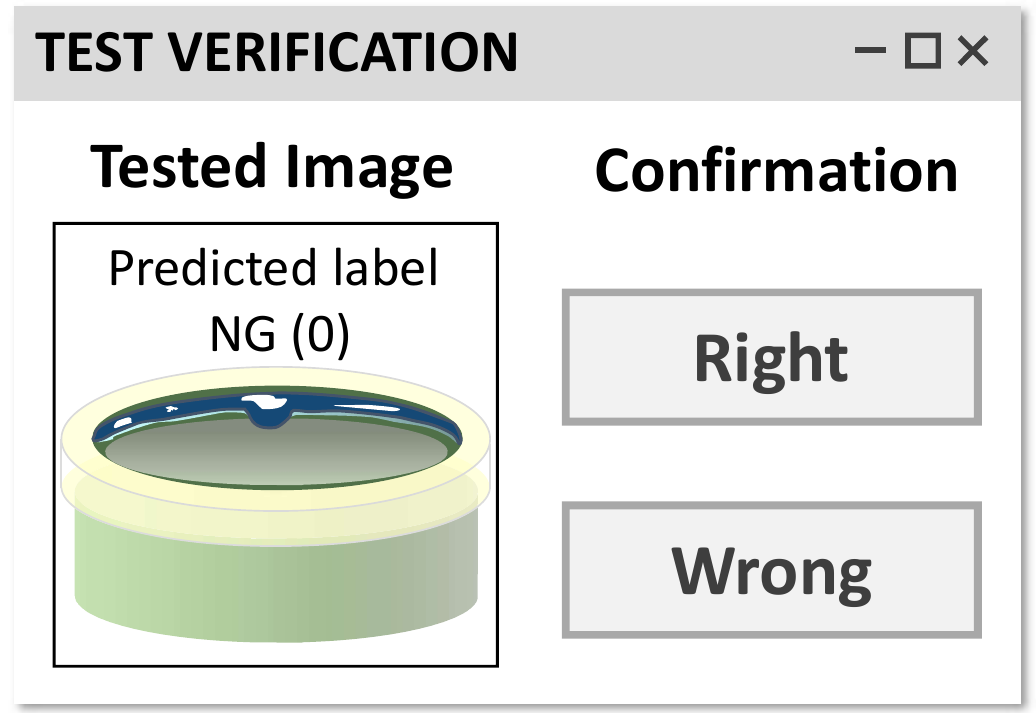}}
		\hspace{5pt}
		\subfigure[Grad-cam verification]{\includegraphics[height=0.32\columnwidth]{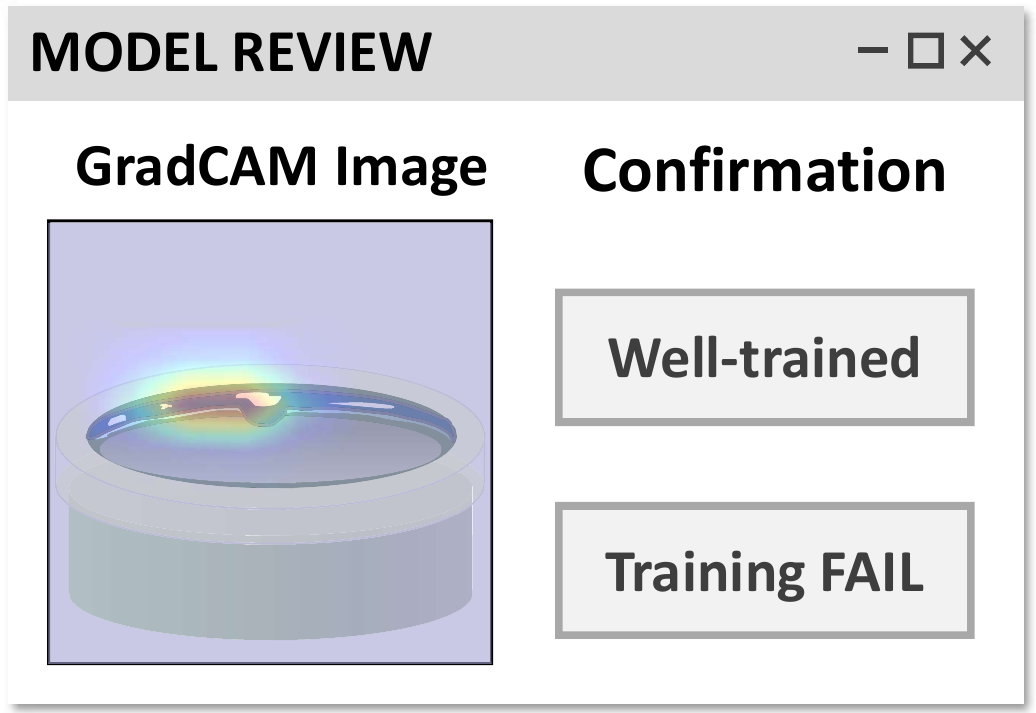}}
		\caption{User interface program for verifying the unreliable test samples.}
		\label{FIG:2stage_test}
	\end{figure}
	
	\subsubsection{Model Update System}
	\label{subsec:model_update}
	
	In general, deep learning models perform better than traditional methods, but they are not permanently perfect.
	For example, the trained model $p\big(y|\mathbf{x};\mathbf{w}\big)$ was perfect for the training dataset $\mathcal{D}$, there is no guarantee that the model always will be perfect for the consecutive testing samples.
	We expect that the distribution of test samples is the same as that of training samples~$\mathcal{D}$. Unfortunately, the distribution of the test samples begins to change subtly over a long period of time due to many factors~(e.g., machines becoming obsolete, raw material of products change and Etc.).
	Therefore, continuous maintenance and model update are required.
	
	In order to update the deep learning model, we can commonly consider two types of update strategies: 1) \textit{re-training} and 2) \textit{fine-tuning}~\cite{pan2009survey}.
	First, \textit{re-training} is to retrain the deep model~(i.e., all weights $\mathbf{w}$) from the beginning using whole training dataset $\mathcal{D}$ and new failed samples. It is reliable, but requires a large amount of computation due to significant training data.
	On the other hand, \textit{fine-tuning}~\cite{pan2009survey} simply adjusts the trained weights $\mathbf{w}$ using only new failed samples. Fine-tuning is very efficient but it is likely to occur a fatal problem so-called \textit{catastrophic forgetting}\footnote{Catastrophic forgetting is a phenomenon in which deep learning models forget previously learned information upon learning new information.}~\cite{french1999catastrophic} when the fine-tuning is repeated too often.
	
	To avoid the challenges in updating deep learning model, we exploit both strategies that complement each other to build an efficient model update system.
	As shown in	Fig.~\ref{FIG:whole_frame}, we first separated main server and edge servers for efficiency and stability. The main server is a high performance workstation~(nvidia tesla V100) and it performs model \textit{re-training}.
	The edge servers equipped `nvidia RTX2080' and they perform fine-tuning tasks for each product inspection system maintenance.
	
	\noindent $\bullet$ \textbf{Failed sample collection.} \quad Assume that a trained classifier $p\big(y|\mathbf{x};\mathbf{w}\big)$ has tested new samples so that we get a set of test samples. Among them, we choose a set of unreliable test samples as follows		
	\begin{equation}
		\begin{split}
			\mathcal{D}^u & =  \big\{ \big(\mathbf{x}_i,y^{pred}_i\big)  | 0.5-\alpha \leq  p\big(y_i|\mathbf{x}_i;\mathbf{w}\big)  \leq 0.5+\alpha \big\}, \\
			& \text{where} \qquad i=1,...,N_t
		\end{split}
	\end{equation}
	where $N_t$ is the total number of tested samples and $y^{pred}$ is a predicted label of $i$-th sample (refer to Eq.(\ref{eq:label})). Note that all testing samples are unseen data ($\mathbf{x}_i\notin\mathcal{D}$).
	We defined the samples with prediction probabilities around $0.5$ as unreliable samples.
	This is because the samples near decision boundary~(0.5) are not clearly distinguished by the classifier $p\big(y|\mathbf{x};\mathbf{w}\big)$.
 	We set $\alpha$ as $0.5$ empirically.
 	
 	There are lots of \textit{uncertainty-based active learning}~\cite{tong2001support} algorithms, and we can change the failed sample collection algorithm with them. However, even with this naive sample collecting logic, we experimentally find out that it confirms robustness.
	
	
	Then, system managers verify the unreliable samples~$\mathcal{D}^u$ through two-stage verification. To this end, we designed a simple WINDOWS application which demonstrates the tested sample images with two buttons as shown in Fig.~\ref{FIG:2stage_test}.
	In the first stage, system managers verify the predicted label of tested sample is right or wrong (Fig.~\ref{FIG:2stage_test} (a)).
	If there are wrong samples, they can build a set of prediction failed samples as
	\begin{equation}
		\mathcal{D}_{f}^{u}  =  \big\{ \big(\mathbf{x}_j,y^{gt}_j\big)  | \mathcal{D}^u, y^{pred}_j  \neq y^{gt}_j \big\}, 
	\end{equation}
	where $j$ is an index of the sample, and $y^{gt}_j$ is a ground-truth label of $j$-th sample.
	Although the predicted label of test sample is right, the system managers verity the sample one more step. As shown in Fig.~\ref{FIG:2stage_test} (b), they check the Grad-cam images of samples to verify the deep learning model properly predicted the label as explained in Sec.~\ref{subsec:exp_system}. According to the result of Grad-cam, we can also build a poorly trained sample set as
	\begin{equation}
		\mathcal{D}_{g}^{u} =  \big\{ \big(\mathbf{x}_j,y^{gt}_j\big)  | \mathcal{D}^u, y^{pred}_j  = y^{gt}_j , G \big(\mathbf{x}_j\big)=0 \big\},
		\label{eq:grad}
	\end{equation}
	where $G\big(\mathbf{x}_j\big)$ denotes a Grad-cam verification result of the sample $\mathbf{x}_j$. If the system manager pushed `Well-trained' button, $G\big(\mathbf{x}_j\big)$ would be `1'. On the other hand, if 'Training FAIL' button was pushed $G\big(\mathbf{x}_j\big)$ is assigned as `0'.
	Thanks to Eq.(\ref{eq:grad}), we will consider the ambiguous samples as well.
	Finally, we get a set of failed samples during the test as
	\begin{equation}
		\overline{\mathcal{D}}=\mathcal{D}_{f}^{u}  \cup \mathcal{D}_{g}^{u}.
	\end{equation}
	It can be represented by $\overline{\mathcal{D}}\big(t\big)$ where $t$ is a time index. E.g., when we set a time slot as $2$ days for collecting failed samples, $\overline{\mathcal{D}}\big(1\big)$ is a failed sample set collected during first and second days. In the same way, $\overline{\mathcal{D}}\big(2\big)$ denotes third and forth days' failed sample set.
	
	\noindent $\bullet$ \textbf{Model update methods.} \quad A current deep model $p\big(y|\mathbf{x};\mathbf{w}\big)$ cannot perfectly classify the samples in $\overline{\mathcal{D}}\big(t\big)$. To make the model more accurate, we need to update the model weights $\mathbf{w}$ based on $\overline{\mathcal{D}}\big(t\big)$.
	We perform \textit{fine-tuning} the model in an edge computer~\footnote{\textit{fine-tuning} should be performed when the production line is idle.} as
	\begin{equation}
		\mathbf{w}^{+}  \leftarrow \mathbf{w}-\mu\frac{\partial  E_{\overline{\mathcal{D}}\big(t\big)}}{\partial \mathbf{w} },
	\end{equation}
	where $\mu$ is a learning rate and $E_{\overline{\mathcal{D}}\big(t\big)}$ is the loss function of the deep model with respect to the dataset $\overline{\mathcal{D}}\big(t\big)$.
	After the \textit{fine-tuning} process, the updated model $p\big(y|\mathbf{x};\mathbf{w}^{+}\big)$ classifies the failed samples and can handle better upcoming test samples. It is simple but very effective in that it requires a small amount of computation and it improves the model performance a lot through a small adjustment in model weight distribution.
	
	However, repeating the \textit{fine-tuning} process too often causes \textit{Catastrophic forgetting}~\cite{french1999catastrophic}. Then the model performance begins to deteriorate since the model fails to classify the samples trained in the past.
	In order to prevent the \textit{Catastrophic forgetting} problem, we periodically perform \textit{re-train} under following conditions.
	First, we measure a failed sample ratio as
	\begin{equation}
		\mathcal{FSR}=\frac{\big|\bigcup_{t=1}{ \overline{\mathcal{ D}}\big(t\big)}\big|}{ \big| \mathcal{D}\big|},
	\end{equation}
	where $\big|\cdot\big|$ is a cardinality of a dataset.
	If the measured $\mathcal{FSR}$ is larger than $\beta$, we update the training dataset as
	\begin{equation}	
		\mathcal{D}^{+} \leftarrow \mathcal{D}  \cup  \big(\bigcup_{t=1}{ \overline{\mathcal{D}}\big(t\big)} \big),
	\end{equation}
	It includes not only original training dataset $\mathcal{D}$ but also newly collected samples $\big(\bigcup_{t=1}{ \overline{\mathcal{D}}\big(t\big)} \big)$ during testing the system.
	Then we \textit{re-train} the model by
	\begin{equation}
		\mathbf{w}^{+}  \leftarrow \mathbf{w}-\mu\frac{\partial  E_{\mathcal{D}_{+} }}{\partial  \mathbf{w} },
	\end{equation}
	where $E_{\mathcal{D}_{+} }$ is the loss function of the deep model with respect to the new dataset $\mathcal{D}_{+}$. The \textit{re-training} process is performed on the main server because it requires a lot of computing resources. When model \textit{re-training} is done, the model weights are transferred to the edge computer. 
	
	\textit{Fine-tuning} and \textit{re-training} processes complement each other and they are repeated continuously during the model maintenance.
	We show several results of proposed model update system in Sec.~\ref{subsec:us}.
	The proposed system is very easy and effective to maintain the model for system managers.

	\begin{table}[t]
		\centering
		\caption{PCB Parts dataset}
		{\small
			\begin{tabular}{c|c|c}
				\noalign{\hrule height 1pt}
				\multicolumn{1}{c|}{Parts type} & \# of defective & \# of non-defective \\ \noalign{\hrule height 1pt}
				soldered pin      & 1,000 & 1,000     \\ \hline
				MCU      & 2,200 & 2,200     \\ \noalign{\hrule height 1pt}
		\end{tabular}}
		\label{Tab:dataset}
	\end{table}

	\section{Datasets}
	\label{sec:DB}
	
	Due to company confidentiality, it is difficult to open the original images of the data.
	Instead, we provide details and illustrations of each dataset.
	According to types of production lines, we have collected \texttt{PCB parts}, \texttt{Cylinder bonding} and \texttt{Navigation icon} datasets as follows:

	\noindent $\bullet$ \texttt{PCB parts} dataset contains two types of components in PCB such as soldered pins~(see Fig.~\ref{FIG:ok_ng_images}) and micro control units~(see Fig.~\ref{FIG:ok_ng_images_mnt}).
	We have collected both defective and non-defective samples for each class as summarized in Table.~\ref{Tab:dataset}.
	
	\noindent $\bullet$ \texttt{Cylinder bonding} dataset contains images of cylinders with bond applied.
	As shown in Fig.~\ref{FIG:gradCAM}, bonds should be evenly applied inside the cylinders unless the  bonding between parts will not work properly. If the bond is broken or lumped together, we consider the cylinder is defective. We have collected 651 non-defective images and 651 defective images.
	
	\noindent $\bullet$ \texttt{Navigation icon} dataset is a set of icon images in the car navigation software. It includes 20 classes of icons. The size of each icon is normalized to $64*64$ pixels. This dataset is insufficient to train deep learning models. In Sec.~\ref{sec:da}, we show some classification results with the \texttt{Navigation icon} dataset.
	
	\section{Experimental Results}
	\label{secar}
	
	\subsection{Data Augmentation}
	\label{sec:da}

	Table.~\ref{Tab:aug_test} shows the results of classification performances before and after the data augmentation process. We tested \texttt{Navigation icon} dataset and classified 20 different types of common icon images in a mobile application. To augment images, we utilized color transformation methods in~\cite{imgaug} and produced ten times more samples.
	As you can see, a classification result after data augmentation improved 22.2\% classification performance.
	It implies that data augmentation is highly beneficial.
	When data is insufficient or unbalanced, we recommend to perform simple data augmentation to handle the lack of data problem.

	\begin{table}[t]
		\centering
		\caption{Performance enhancement through data augmentation.}
		{\small
			\begin{tabular}{c|l|c}
				\noalign{\hrule height 1pt}
				Class & \multicolumn{1}{c|}{The number of images} & Accuracy. \\ \noalign{\hrule height 1pt}
				20    & 272 (Original data)                        & 0.632\%     \\ \hline
				20    & 2720 (Original data + Augmented data)      & 0.854\%     \\ \noalign{\hrule height 1pt}
		\end{tabular}}
		\label{Tab:aug_test}
	\end{table}

	\subsection{Defect Classification Results}
	\label{subsecar 4.1}
	
	In this section, we address experimental results on defect classifications in terms of datasets: \texttt{PCB parts}, \texttt{Cylinder bonding}.
	
    \subsubsection{\texttt{PCB parts} dataset}
    \label{subsubsec:pcb}
    By training the classification networks, we first tested the classification performance for the soldered pins. To find the model with best performance, a total of five based network were used such as Resnet~\cite{he2016deep}, vgg16~\cite{simonyan2014very}, GoogleNet~\cite{szegedy2015going}. These models were trained in same environment (e.g., training and test data set, hyperparameter etc.). As a result, Resnet~\cite{he2016deep} showed the highest accuracy for classification of the soldered pins (see Table.~\ref{TAB:solder}). Based on the results, we used Resnet~\cite{he2016deep} as the basic model when verifying the performance of the other data sets.
	
	On the other hand, to find defective MCU in \texttt{PCB parts} dataset, we trained defect detection model based on Yolo v3~\cite{redmon2018yolov3}. 
	Figure~\ref{FIG:ok_ng_images_mnt} shows examples of defects on the MCU part, and the defects occur partially.
	As we studied in Sec.~\ref{sec:ss3}, utilizing defect detection model is very effective when the defects occur partially.
	Table.~\ref{TAB:mcu} shows performance comparisons between classification-based models and detection-based model. Yolo v3~\cite{redmon2018yolov3} showed the highest performance to detect defective MCU samples.
	On the other hand, classification-based model which is used for classifying the soldered pins often failed to detect partial defects on MCU parts.
	Based on these experimental results, we verified proper deep learning models (e.g., defect classifier, defect detector) for each dataset.
	
	\subsubsection{\texttt{Cylinder bonding} dataset}
	
	Adhesive bonding is a process where the classification algorithm can be applied. Bond is applied to a part to combine two parts, and mostly a vision inspection is carried out in this process. The errors that occur in this process include over-gluing, insufficient gluing, air bubbles, and slipping~\cite{haniff2011shape}. The classification algorithm is more favorable to this process than the detection algorithm unless the bonding area is wide. This is not only because the inspection duration of classification is normally shorter than detection, but also it is important to see the overall shape of a part for certain types of defects.
	
	As we tested in Sec.~\ref{subsubsec:pcb}, ResNet~\cite{he2016deep} shows best performance for defect classification. Therefore, we employ the deep learning model:ResNet~\cite{he2016deep} as the baseline model for this experiment.
	In addition, We made comparisons among the applicable, latest algorithms of ResNeXt-50, Se-ResNet-50, Se-ResNeXt-50 while taking CT of the process into consideration. For the used images, we used 20 defective, 20 non-defective for learning, and 110 defective, 110 non-defective for testing, and changed the image size to 224x224x3 before comparison. For the reliability of the test performance assessment, we conducted 5 rounds of random verification on accuracy and loss. The results are summarized in Table~\ref{TAB:cylinder}.
	
	The reason for the result is assumed that as this requires to focus on certain defect areas instead of seeing the overall shape, Se-ResNet-50 with attention lines showed a better performance than ResNeXt-50.

	\begin{table}[]
		\centering
		{\small
			\caption{Performance comparison of \texttt{Cylinder bonding}}
			\label{TAB:cylinder}
			\begin{tabular}{l|c|c|c}
				\noalign{\hrule height 1pt}
				Method        & Avg.Loss & Mean Acc. & GFLOPs \\ \noalign{\hrule height 1pt}
				ResNet-50~\cite{he2016deep}     & 0.28260  & 0.941\%   &  3.86  \\ \hline
				ResNeXt-50~\cite{xie2017aggregated}    & 0.58632  & 0.915\%   &  4.24  \\ \hline
				Se-ResNet-50~\cite{hu2018squeeze}  & 0.17950  & 0.969\%   &  3.87  \\ \hline
				Se-ResNeXt-50~\cite{hu2018squeeze} & 0.12076  & 0.975\%   &  4.25  \\ \noalign{\hrule height 1pt}
			\end{tabular}
		}
	\end{table}
	
	\begin{table*}[]
		\centering
		{\small
			\caption{Performance comparison of \texttt{PCB parts}: soldered pin dataset}
			\label{TAB:solder}
			\begin{tabular}{c|c|c|c|c|c}
				\hline
				Methods  & Xception~\cite{chollet2017xception} & Resnet-50~\cite{he2016deep} & vgg16~\cite{simonyan2014very} & vgg19~\cite{simonyan2014very} & GoogleNet~\cite{szegedy2015going}                               \\ \hline
				Accuracy & 0.954     & \textbf{0.985}  & 0.954  & 0.972  & 0.963       \\ \hline
			\end{tabular}
		}
	\end{table*}

	\begin{table*}[]
		\centering
		{\small
			\caption{Performance comparison of \texttt{PCB parts}: MCU dataset}
			\label{TAB:mcu}
			\begin{tabular}{c|c|c|c|c|c|c}
				\hline
				& \multicolumn{5}{c|}{Classification-based}     & \multicolumn{1}{l}{Detection-based} \\ \hline
				Methods  & Xception~\cite{chollet2017xception} & Resnet-50~\cite{he2016deep} & vgg16~\cite{simonyan2014very} & vgg19~\cite{simonyan2014very} & GoogleNet~\cite{szegedy2015going} & Yolo v3~\cite{redmon2018yolov3}                               \\ \hline
				Accuracy & 0.89     & 0.83   & 0.90  & 0.85  & 0.85      & \textbf{0.998}                                 \\ \hline
			\end{tabular}
		}
	\end{table*}

	\begin{figure}[t]
		\centering
		\subfigure[]{\includegraphics[width=1\columnwidth]{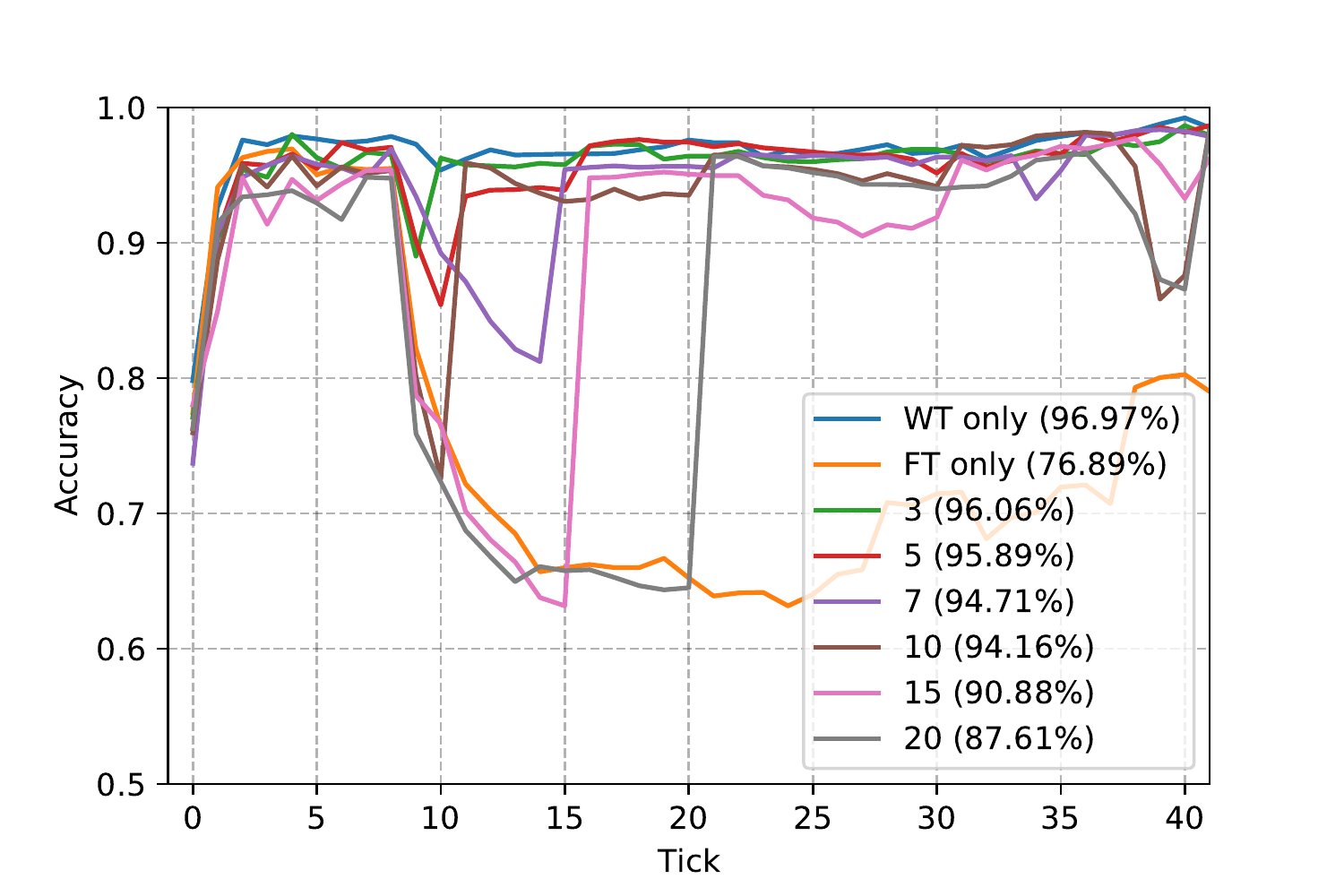}}
		\subfigure[]{\includegraphics[width=1\columnwidth]{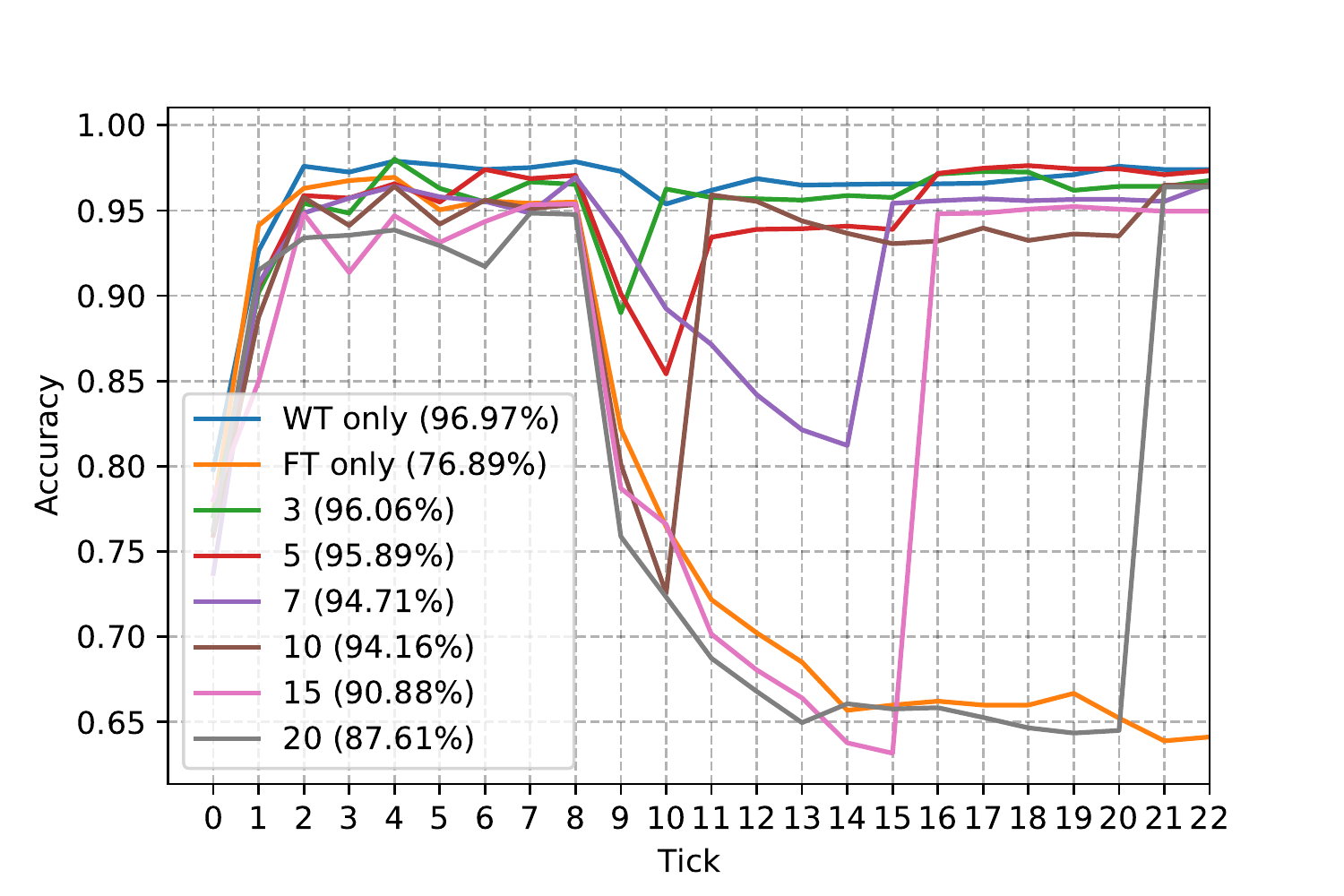}}
		\caption{Model accuracy according to update ticks. Will be modified.}
		\label{FIG:tuning}
	\end{figure}

	\subsection{User systems}
	\label{subsec:us}
	We first designed an experiment to obtain an experimentally appropriate $\mathcal{FSR}$ rate. The experiment was conducted with cylinder bonding images collected over the past two years. The initial training was done using 60 images (good:30, defective:30) collected in the first two months, and the test accuracy was calculated using 262 images (good:131, defective:131) also obtained within the first two months that were not used for training. It was assumed that for every 15 days(1 tick in Figure~\ref{FIG:tuning}), 20 additional images (good:10, defective:10) were obtained and used for training. Figure~\ref{FIG:tuning} is the accuracy comparison graph for the existing image according to the number of fine-tuning. We also set \textit{fine-tuning parameters} to find the right $\mathcal{FSR}$ rate, which means the frequency of re-training with entire data. For example , 3 \textit{fine-tuning parameters} means that 1 whole re-training is performed every 3 fine-tuning, and \textit{WT only} means Whole re-training only and \textit{FT only} means fine tuning only.
	
	As you can see in the graph, there are some dramatic deteriorating points with accuracy under 90\%. When the accuracy is below the expected percentage, it is considered inaccurate, as the system failed to recognize previous images. Thus, it is considered that the forgetting problem is happened. For example, 11 tick with 5 \textit{fine-tuning parameters} and 40 tick with 10 \textit{fine-tuning parameters}, $\mathcal{FSR}$ rate of both points are 0.375 and 0.220. Forgetting problem does not occur with the entire points of less then 0.2 $\mathcal{FSR}$ rate. When performing fine-tuning, it has been verified experimentally that the full re-training must be conducted with less then 0.2 $\mathcal{FSR}$ rate to prevent forgetting problem. and we applied it to the system, if $\mathcal{FSR}$ rate reach to 0.2 the server will do the full re-training. 
	In the experiment, we also confirmed that as $\big| \mathcal{D}\big|$ increases, the model becomes much robuster, so that the training \textit{loss} for $\big|\bigcup_{t=1}{ \overline{\mathcal{ D}}\big(t\big)}\big|$ decreases, so even if a large number of $\big|\bigcup_{t=1}{ \overline{\mathcal{ D}}\big(t\big)}\big|$ are newly trained, the deterioration of the model performance appears more lately.

	\section{Conclusions}
	\label{sec:conclusion} 
	
	In this work, we proposed a unified framework for product quality inspection using deep learning techniques.
	We categorized several deep learning models that can be applied to product inspection systems. In addition, we studied which deep models are proper for   
	The entire steps for building a proposed framework for product inspection system via deep learning are explained in details.
	In addition, we addresses several connection schemes for linking the deep learning models to the existing product inspection systems.
	Finally, we proposed effective model managing methods that efficiently maintain and enhance the deep learning models. 
	It showed good system maintenance and stability.
	
	We tested and compared performance of state-of-the-art methods to verify the effectiveness of the proposed methods in various test scenarios.
	We expect that our studies will be helpful and will give a great guidance for those who want to apply deep learning techniques on product inspection systems.


	\appendix
	\section{Appendix}
	\label{Append1}

	\subsection{Product Manufacturing Systems}
	\label{sec:MES_PLC}
	
	A defect detection system is a system running in connection with various process systems. To apply deep learning to the defect detection system, a comprehensive understanding of various process systems is required. Although there are many different systems according to the size or systems of a factory, below are the key systems relevant to our research.
	
	\begin{itemize}
		\item Manufacturing Execution System (MES)~\cite{kletti2007manufacturing}
		\item Programmable Logic Controller (PLC)~\cite{reis1998programmable}
	\end{itemize}
	
	MES~\cite{kletti2007manufacturing} is an information system that helps decision making by storing and providing multiple data generated in a series of production activities from the first to the last process of a planned product. The results of quality inspections of each process will also be saved in MES. Therefore, to apply deep learning to production processes, the connection between a deep learning system and MES is needed.
	
	PLC~\cite{reis1998programmable} refers to a control device or system that makes input/output of equipment operate according to a certain sequence. The product inspection results will be closely intertwined with the next process of the product and be controlled by PLC. For example, products turned out to be defective will be piled up on a buffer for scrap or repair, and non-defective products will be sent to the next process (assembly or packaging) by PLC.

	{\small
		\bibliographystyle{ieee}
		\bibliography{egbib}
	}

\end{document}